# Adapting Vision Transformer for Efficient Change Detection

Yang Zhao, Yuxiang Zhang, Yanni Dong, *Senior Member, IEEE,* and Bo Du, *Senior Member, IEEE*

*Abstract*—Most change detection models based on vision transformers currently follow a *"pretraining then fine-tuning"* strategy. This involves initializing the model weights using large-scale classification datasets, which can be either natural images or remote sensing images. However, fully tuning such a model requires significant time and resources. In this paper, we propose an efficient tuning approach that involves freezing the parameters of the pretrained image encoder and introducing additional training parameters. Through this approach, we have achieved competitive or even better results while maintaining extremely low resource consumption across six change detection benchmarks. For example, training time on LEVIR-CD, a change detection benchmark, is only half an hour with 9 GB memory usage, which could be very convenient for most researchers. Additionally, the decoupled tuning framework can be extended to any pretrained model for semantic change detection and multi-temporal change detection as well. We hope that our proposed approach will serve as a part of foundational model to inspire more unified training approaches on change detection in the future. Code will be available at URL.

## I. Introduction

Change Detection (CD) is one of the fundamental tasks in earth observation [1]. By interpreting images of the same region from different periods, we can understand the economic development and human activities via observing changes such as building demolition and farmland conversion. For example, Microsoft researchers employed images before and after geological disasters for change detection to evaluate the post-disaster damage of large-scale artificial buildings in Turkey, providing assistance for rescue effort. This work is a typical example applying deep learning method to extract spatial and temporal features for change detection. Generally, most deep learning-based change detection methods [2], [3] are based on convolution neural network. The first deep learning-based change detection method FC-Net [4] based on U-Net [5], originally developed for medical applications, has the ability to capture small changes and is an important baseline. Its three fusion methods including early concatenation, late concatenation, and late difference, have also been incorporated into newer models. For example, [6] and [7] use the late difference as their fusion operation by comparing the output of two corresponding layers and calculating the difference.

Transformers [8], [9] are a popular topic in architecture across various media, including image [8], video [10], [11], text [12] and audio. This is due to their ability to model globally and act as unified feature extractors. Recently, many transformer-based methods [6], [7], [13], [14] have shown promising results in change detection benchmarks. However, the success of these methods heavily relies on large datasets due to their lack of image bias, compared to convolutional neural networks (CNNs) [15]. Unfortunately, change detection task suffers from a lack of sufficient training samples. This results in training transformers from scratch taking longer to converge and being more prone to overfitting. Therefore, vision transformer based change detection methods typically load pretrained model parameters before starting the target task's training process. Although loading pretrained models can lead to better performance, it requires a significant investment of time and resources due to the large size of these models. For example, the training in [16] on a single Tesla RTX8000 GPU can take up to more than 40 hours. The large models including transformers have shown strong generalization ability not only in change detection [4] but also in natural language processing (NLP) [12] and computer vision domains [13], albeit at the expense of expensive training time and resources.

To tackle the issues of training consumption, a new research direction called parameter-efficient transfer learning [17]–[22] has emerged in the field of NLP. This research aims to achieve optimal performance by only fine-tuning a small number of additional parameters, while keeping large pretrained language models unchanged. Such techniques have been recently introduced to image classification [23], [24] (image-to-image) and video understanding [25] (image-to-video) in computer vision. First of all, under this parameter-efficient transfer learning paradigm, all downstream tasks share a set of parameters for the pretrained model. This requires the pretrained model to have extensive feature extraction capabilities, which has recently been addressed by self-supervised learning [26]–[29] of computer vision. Secondly, a framework that can specifically solve downstream tasks and keep the backbone model consistent with the upstream model is needed. To our knowledge, directly leveraging a pretrained model for change detection tasks (image-to-temporal image) is less explored, which is likely due to the absence of a standardized change detection framework for transferring

This work was supported in part by the National Natural Science Foundation of China under Grants 62071438, 62222116, 62225113, and 62171417. *(Corresponding author: Yuxiang Zhang)*

Y. Zhang, Y. Zhao, and Y. Dong are with School of Geophysics and Geomatics, China University of Geosciences, Wuhan 430074, China (e-mail: zhangyx@cug.edu.cn; zyang6194@gmail.com; dongyanni@cug.edu.cn).

Bo Du is with the School of Computer Science, Wuhan University, Wuhan 430079, China (e-mail: gunspace@163.com).

image models as well as the underwhelming performance of the existing frameworks.

Inspired by the above ideas, we propose a unified framework for change detection that involves adapting the parameters of a pretrained vision transformer model by training a small number of additional layers. Behind of the pretrained vision transformer, we add our fusion module to detect changes between two temporal features and an upsampling module to complete the framework. The inserted additional layer consists of two fully connected layers and a non-linear activation function. This approach significantly reduces training time and resource costs while achieving competitive or even superior results compared to state-of-the-art methods. Furthermore, we compared several mainstream parameter-efficient transfer learning methods [18], [21], [22], [24] for their effectiveness in change detection. Our findings suggest that a well-pretrained natural image model is sufficient for spatial modeling in change detection tasks. Additionally, we discover that early temporal modeling in the feature encoder is unnecessary for the current two-temporal change detection task. To summarize, we make the following contributions:

1) We propose a new way to adapt pretrained image transformer models for efficient change detection. Our Adapter-CD method is a general, simple, and cost-effective solution for adapting different image pretrained models.
2) Our method is significantly more efficient than full fine-tuning a change detection model, e.g., on DINOv2 backbone with the flash attention, training time can be reduced to half an hour and memory footprint can be decreased to 9GB. In contrast, it takes more than 40 hours to train on a single RTX 8000 with a memory capacity of 48G in ChangeFormer [16].
3) We compare the effectiveness of mainstream parameter efficient fine-tuning methods in change detection. We found that fine-tuning a small number of parameters on the natural image model can achieve good results in change detection.
4) Our method also achieves the state-of-the-arts on four change detection benchmarks with lower tunable parameters.

## II. RELATED WORK

### A. Transformer in Vision Field

Transformer [9], which has unified the model structure of natural language processing, has gradually demonstrated its dominance in the field of vision in recent years, especially in tasks that require advanced semantic expression. In some dense prediction tasks, transformers [30], [31] have achieved impressive results even without hierarchical feature extraction. The hierarchical feature can be utilized for subsequent operations to achieve the fusion of spatial and advanced semantic information, which can improve the accuracy of semantic segmentation and change detection. However, it is not mandatory to employ hierarchical feature extraction while loading the pretrained transformer model for some dense prediction tasks. One possible reason is that self-supervised learning particularly with transformer not only benefits semantic extraction but also enhances the extraction of multi-level information. Transformer has been applied in various vision tasks, including image classification [8], [32], object detection [33], instance segmentation [30], video understanding[10], [11], and change detection [6], [13], [16].

Due to its extensive fitting and generalization capabilities, transformer can be combined with self-supervised learning to learn more general features from large-scale data [28], [29], [34], [35]. In this case, the large data dependency issue of the transformer can be resolved and its accuracy can also be improved. Then, the trained model could also serve as good initialization for transfer learning to downstream tasks [31]. Therefore, in nowadays, loading a pretrained transformer model to achieve better performance especially for tasks with insufficient data is very common. Additionally, there exists more than thousands of pretrained transformer models in the web, e.g., Huggingface [36] and Timm [37] . By fully tapping into the generalization ability of these pretrained models in some downstream tasks, it can reduce the computing resources consumption and the greenhouse gas emission. In this work, we aim to take advantage of these well pretrained models and adapt them efficiently in change detection.

### B. Interaction with Temporal Images

The integration of temporal images was first proposed in video understanding [38], where the feature dimensions of multi-temporal images were concatenated. There are three versions of concatenation based on the time of fusing image: Early Fusion (fusing the input data), Slow Fusion (fusing temporal information throughout the network), and Late Fusion (fusing the features after the network). This concatenation operation has inspired the development of the first change detection method based on deep learning, known as FC-Net [4]. Furthermore, the difference operation is utilized in certain methods when identifying the type of change in an object is the objective of change detection. Several methods [6], [16] have been developed based on this difference operation to enhance feature extraction by utilizing a more robust spatial encoder. Recently, [39] proposes a new temporal fusion operation that involves exchanging feature dimension and temporal dimension between two input datasets.

Due to the popularity of transformer, some works [10], [11] have also employed self-attention for temporal fusion modeling. However, training self-attention directly on patches across multiple frames can be time-consuming and resource-intensive. A more effective approach is to model corresponding spatial positions across frames in order to integrate temporal information. The most relevant work to us is [40], which involves identifying and selecting objects that have changed in two temporal natural images by using cross-attention. To clearly explain the specific fusion method, let's take the first image as an example, which is identical to the second image. This method first uses the first image as a query and the second image as key and value. Query, key, and value are linear transformations of input data. Then, the method will perform cross-attention calculation to retrieve features related to the first image from querying the second image. The weighted result will then be concatenated with the first image's features to obtain the first feature map containing information from both images, thus enabling identification of changes in location.

By introducing the above fusion method, we can summarize that all of the above methods operate on two input data. In our work, we use masked cross-attention to detect changes between two temporal images, and it can be extended to inputs with multiple temporal data while maintaining low computation. More information is included in section III.A.

*C. Parameter efficient fine-tuning*

Since the appearance of BERT [12] with 350 million parameters, models are becoming larger in natural language processing while computational resources are increasing less than 10 times due to the high cost of high bandwidth memory (HBM) memory. In the vision field, the parameters of vision transformer [41] are extended to 22 billion. And Meta proposes a large self-supervised learning method, namely DINOv2 [28], which has 1.1 billion parameters. The above large models have achieved good performance in various downstream tasks. However, it makes fine-tuning such a large model to downstream tasks infeasible for most researchers. To solve the above problem, parameters efficient fine-tuning (PEFT) [19] was first proposed in natural language processing. PEFT only trains part of existing backbone or newly-added parameters, resulting in a more efficient training process with a smaller memory footprint and time consumption. Even though most parameters are frozen during the training process, the performance of the model is comparable or even superior to that achieved through full fine-tuning. Recently, parameter efficient transfer learning is also studied in the vision field [23]–[25]. They mainly focus on in-domain transfer learning, e.g., natural image classification. To our best knowledge, we are the first to adapt pretrained image model to remote sensing image, especially for change detection task. Additionally, we conduct a comparison of the main parameter efficient fine-tuning methods on change detection. Furthermore, our method is compatible with different image models, and we fine-tuned the state-of-the-art in self-supervised learning, DINOv2 [28]. Our results achieve competitive or state-of-the-art performance on six change detection benchmark datasets.

## III. METHODOLOGY

In this section, we first describe the pretraining model vision transformer (ViT) [8] in section III.A and then introduce the unified change detection framework in section III.B that could efficiently transfer the pretrained image model. Finally, we interpret the adapter layers to change detection in section III.C, which can adapt the parameters of the pretrained model.

*A. Preliminary*

Since vision transformer was developed from natural language processing, transformer-based models have been widely adopted in various remote sensing tasks, such as change detection, hyperspectral classification. In our work, we use the original vision transformer as the backbone of the change detection framework allowing for compatibility with a wider range of pretrained models. This method differs from existing transformer-based changed detection methods that primarily focus on multi-scale feature extraction. Additionally, this method enables the framework to potentially handle more complex change detection tasks involving multimodal tasks that incorporate adding text or elevation images.

Given the input image $x \subseteq R^{3 \times H \times W}$, vision transformer first splits it into $N$ non-overlap patches. Here, $H$ and $W$ are the height and width of the input image respectively. We denote the patch size as $P$ and the number of patches is $N = \frac{H \times W}{P^2}$. Then, a linear layer called patch embedding is operated on these patches, resulting in output token $x_p \in R^{N \times D}$, where $D$ is feature dimension, usually 384 or 768. Then, a learnable class token is prepended to $x_p$ as $x_0 = [x_{class}; x_p] \in R^{(N+1) \times D}$ and $x_{class}$ can be used as the output of the final global feature. Due to the loss of position information when adapting the 2D image to 1D sequences, a learnable position embedding $E_{pos} \in R^{(N+1) \times D}$ are added to $x_0$ as $z_0 = x_0 + E_{pos}$. At this point, the total operations of dividing the image into sequences is complete. Then $z_0$ is input to the sequences of transformer layers to do self-attention. Each transformer layer is composed of multi-head self-attention layer (MSA), multilayer perception (MLP), layernorm (LN), and residual layer. The computation is listed as following.

$$z_l = z_{l-1} + MSA(LN^1(z_{l-1})) \quad (1)$$
$$z_l = z_l + MLP(LN^2(z_l)) \quad (2)$$

Here, $l$ and $l-1$ are the indexes of transformer layer where $z_0$ is the first input to transformer layer. The number of layers is 12 in both the small and base versions of the vision transformer, which is also our default choice. Class token $x_{class}$ is removed before the output feature when the downstream tasks are some dense prediction tasks.

*B. Change Detection Framework*

Except for spatial modeling on single image, temporal fusion method is the most important for change detection. Some traditional temporal fusion operation is introduced in section II.B. Our fusion method is inspired by masked cross-attention. Masked cross-attention (MCA) was initially proposed for predicting the next word in a single sentence in natural language processing [9]. It was then applied to object detection [33] in the vision field, leading to the development of the first end-to-end transformer object detection model. This success prompts masked cross-attention as a unified detection head neural network for semantic and instance detection [42]. SAM [43], the first fundamental vision model, has implemented masked cross-attention for object segmentation based on input dot, box, and even text. The methods mentioned above mainly detect objects in spatial features, whereas our approach focus on detecting object changes in the temporal dimension. We achieve this by implementing masked cross-attention for identify changes between two temporal images. Our method can potentially scale to three or more temporal images and we have yet to explore this possibility in this work.

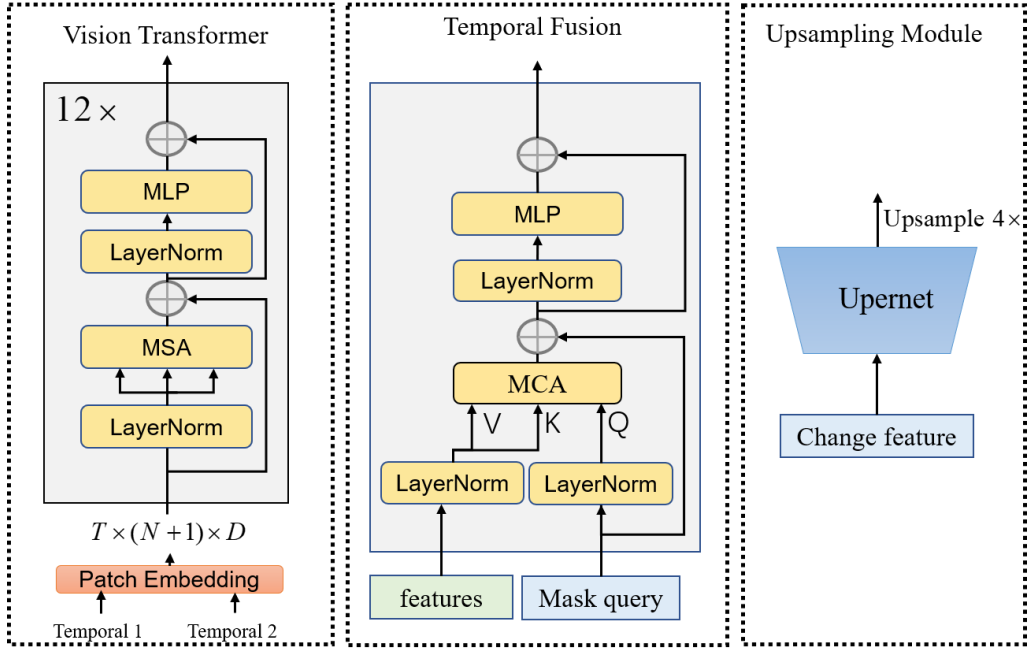

Fig. 1. Change detection framework.

Given the embedding $z \in R^{T \times N \times D}$ from the output of backbone, we first reshape it to $z^T \in R^{N \times T \times D}$, where $T$ is the number of frames, usually 2, and $N = \frac{H \times W}{P^2}$ is the number of patches because the class token $x_{class}$ is removed. Then, the learned temporal position embeddings $z_{TP} \in R^{N \times 2 \times D}$ are added to $z^T$. Finally, we randomly initialize a token, namely mask query $z_m \in R^{N \times 1 \times D}$ to do interaction with the output features $z^T$ in the temporal dimension. To implement masked cross-attention, we first do the query transformation on $z_m$ and separately do the key and value transformation on $z^T$. Here, $query$, $key$, $value$ are all linear layer. Then, the obtained query, key, and value will do scale dot attention computation. The masked cross-attention is computed as follows.

$$Q = query(z_m); K = key(z^T); V = value(z^T) \quad (3)$$

$$O = out(\text{softmax}(\frac{QK^T}{s})V) \quad (4)$$

Here, $out$ is another linear layer to transform the feature dimension to original dimension and $s = \sqrt{D}$ is the scale. Furthermore, we also add norm layer, skip connections, and feed forward layer in the masked cross-attention (MCA) block. Therefore, the complete calculation is as follows.

$$z^T = tranpose(z) \quad (5)$$
$$z^T = z^T + z_{TP} \quad (6)$$
$$z^T = LN^1(z^T); z_m = LN^2(z_m) \quad (7)$$
$$z_m = z_m + MCA(z_m, z^T) \quad (8)$$
$$z_m = z_m + MLP(LN^3(z_m)) \quad (9)$$

The complete pipeline of the change detection framework is shown in Fig 1. The two temporal images $x_1, x_2 \in R^{3 \times H \times W}$ are firstly concatenated in temporal dimension, resulting in $x_T \in R^{T \times 3 \times H \times W}$, where $T$ is commonly 2 and can also scale to more temporal image inputs as stated in former paragraph. Secondly, temporal image $x_T$ is forwarded to the vision transformer, outputting spatial features $x_S \in R^{T \times C \times H \times W}$. Thirdly, we add our MCA block behind vision transformer to do temporal fusion modeling. Finally, we use UperNet [44], commonly used in change detection and semantic segmentation, as our last module to upsample the change feature to original input image size. Compared to UperNet

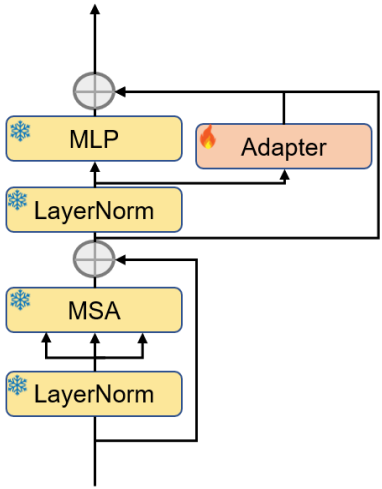 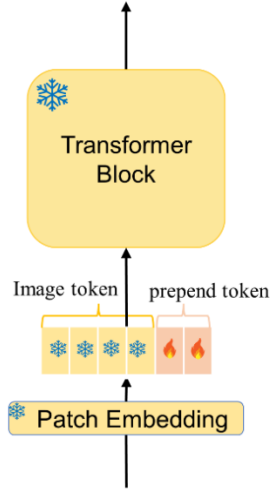 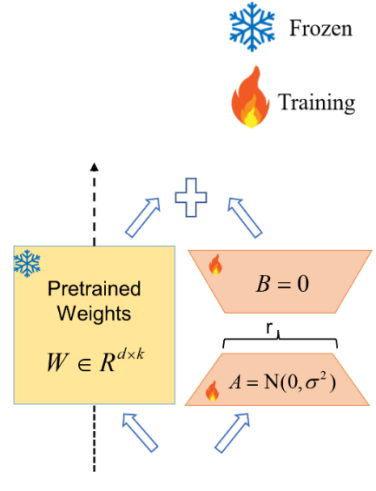

Fig. 2. Adapter layer.   Fig. 3. Prefix-tuning.   Fig. 4. Lora.

[44], we notice that MaskFormer [42] has been utilized in several state-of-the-art methods for other dense prediction tasks. However, its computation and resource requirements are expensive. Additionally, most change detection benchmarks only involve two classes while the advantage of MaskFormer is to deal with multi-class classification.

*C. Adapter Layer*

Even though pretrained image models trained from large scale natural dataset have shown strong transferability, using the encoded feature directly for remote sensing change detection can lead to sub-optimal performance due to the variability in tasks and data domains. Recently, [24] adopted an parameter efficient fine-tuning method in the vision filed and closed the gap with full tuning in terms of accuracy. Therefore, we adopt this adapter layer method to change detection due to its simplicity and good performance. In addition, we also evaluate the efficacy of several other parameter efficient fine-tuning techniques that have been successfully implemented in the domains of video understanding and image generation. Specifically, we compare the performance of three such methods, prefix-tuning [21], Lora [18] and IA3 [22], when employed for change detection task in our experiment.

As stated in section III.A, image is split into many non-overlap tokens before inputting to the transformer layer. Prefix-tuning prepends some learnable tokens to these image tokens, the output features can be aligned to target task. And the learnable token can be prepended to any transformer layers. Lora hypothesizes that change of weights are low rank when tuning the pretrained model for downstream task. Then, a low rank dense layer is inserted into self-attention layer as a substitute for updating the parameters of the model during transfer learning, while keeping the pretrained model parameters unchanged. $W \in R^{d \times k}$, the low rank layer composed of weight matrix $A \in R^{d \times r}$ and weight matrix $B \in R^{r \times k}$ is adopted to replace the behavior of updating the pretrained weights $W$. Among them, $A$ uses Gaussian initialization and $B$ uses zero initialization. This can ignore the newly inserted parameters in the early stage of training and reduce the instability. IA3 can be seen as a special export form of Lora, and its direct purpose is to control the scale of pretrained weights. IA3 sets the low rank dense layer as a single matrix $A$ with rank one, and multiplies the output of pretrained weights and new inserted layers element-wise. Because the inserted layer has low rank and the most parameters of the model are frozen in the above two methods, the computation is very low. And it is very hotly used for model fine-tuning in the field of image generation. The sketch maps of prefix-tuning and Lora are shown in Fig. 3 and Fig. 4.

The adapter layer is an effective tuning method that has demonstrated superior performance in image classification [21]. As a result, we choose it as our default option. Adapter layer is composed of a down-projection layer, an activate function, and an up-projection layer, as shown in Fig. 2. Experiments in several papers [20], [24] have found that inserting such a bottleneck layer after feed forward layer in parallel is the most effective. In the early stage of training, the inserted layers do not disturb the output feature when initializing to zero. To be more precise, scale $s$ is used to weight the feature from the adapter layer. Given the feature $z$ from the output of the self-attention, the computations are as follows.

$$z = z + MLP(LN(z)) + s \times W_{up}(GELU(W_{down}(z))) \quad (10)$$

Here, $MLP$, $LN$, $GELU$ are feed forward layer, LayerNorm, and activate function, respectively. $W_{up} \in R^{D' \times D}$ and $W_{down} \in R^{D \times D'}$ are up-projection layer and down-projection layer, where $D'$ is bottleneck dim and $D' \ll D$. As similar to

Lora, most parameters of pretrained model are frozen.

## IV. EXPERIMENT

In our experiment, we fist evaluate the effectiveness of adapting image model with linear-prob and some other parameter efficient fine-tuning methods on remote sensing change detection. Then we evaluate our strategy with state-of-the-art change detection methods on six change detection benchmarks. Finally, we conduct experiments to ablate the pretrained model, model size and hyperparameters about some tuning methods. We introduce the six change detection benchmarks in section IV.A, training details in section IV.B, the evaluation of efficient tuning methods in section IV.C, comparison with SOTA methods in section IV.D and more ablation studies in section IV.E.

### A. Datasets

*S2looking* [45]: a building change detection dataset containing large-scale side-view satellite images taken at different off-ground angles. The dataset comprises 5000 registered bit-time image pairs with a size of 1024×1024 and a resolution ranging from 0.5 to 0.8 m/pixel. Additionally, it contains over 65,920 annotated instances of change in rural areas worldwide that are classified into three categories: newly constructed buildings, demolished buildings, and background.

*LEVIR-CD* [46]: this change detection dataset consists of 637 pairs of Google Earth (GE) image patches, each with a high resolution of 0.5 m/pixel and a size of 1024×1024 pixels. These images were captured from 20 different regions in various cities across Texas, USA between the years 2002 and 2018. The dataset includes various types of buildings such as villa residences, high-rise apartments, small garages, and large warehouses.

*LEVIR-CD+* [45]: The dataset is a continuation of the LEVIR-CD dataset and has over 985 bitemporal Google Earth images with a very high resolution of 0.5m/pixel. The images are 1024×1024 pixels in size and were collected from 20 different regions in various Texas cities, spanning from 2002 to 2020.

*DSIFN* [47]: this dataset for change detection consists of high-resolution bi-temporal images that were manually collected from Google Earth. It covers six cities in China, including Beijing, Chengdu, Shenzhen, Chongqing and Wuhan. Five of the large image-pairs have been divided into 394 sub-image pairs with dimensions of 512×512 pixels. The dataset includes a total of 3600 image pairs for training, 340 for validation and 48 for testing purposes.

*WHU-CD* [48]: this dataset includes an area that experienced a 6.3-magnitude earthquake in February 2011 and was subsequently rebuilt over the following years. It comprises aerial images captured in April 2012, depicting 12,796 buildings across a span of 20.5 km² (compared to the 16,077 buildings present in the same area as depicted by the 2016 dataset).

*CDD* [49]: This dataset contains 7 pairs of season-varying images with a resolution of 4725×2700 pixels. It also includes 4 pairs of season-varying images with minimal changes and a resolution of 1900×1000 pixels, which were used to manually add additional objects. The dataset was generated by cropping the images to a size of 256×256 pixels. The dataset consists of 16,000 image sets with an image size of 256×256 pixels: 10,000 for training and 3,000 each for testing and validation.

### B. Training details

We use PyTorch and PyTorch Lightning to build model training framework. And the pipeline part of data preprocessing is modified from TorchGeo. For the default settings of all experiments, we use a batch size of 32, 100 number of epochs, AdamW optimizer, weight decay of 0.05, and betas of (0.9, 0.999). We set learning rate of 0.0004, patch sizes of 512×512, and a training/validation split of 0.2 for datasets which did not have a predefined validation split based on baseline model for the dataset. In dataset preprocessing, training samples are all random cropped by 512, and augmentation contains only horizontal flip and vertical Flip. We use the combined loss of focal and Jaccard as the loss function. Specially, for the WHU-CD, LEVIR-CD, and CDD dataset, we crop image to 256 following ChangeFormer.

### C. Parameter Tuning methods on change detection

This section evaluates the effectiveness of the proposed adapter on three datasets. The experiments compare the adapter, other parameter efficient fine-tuning methods and linear-prob that training just a linear classifier on WHU-CD, LEVIR-CD+, and CDD change detection datasets. We use the DINO as our pretrained model, which is the first framework to use a vision transformer for self-supervised learning and achieves better performance compared with other self-supervised learning methods. Results on different pretrained weights are shown in section IV.E. The backbone is small version of vision transformer and we report evaluation metrics on test dataset. The F1 score (F1), Intersection over Union (IoU), and Overall Accuracy (OA) performance of compassion parameter tuning methods on three datasets are illustrated in Table I. The best values of each evaluation metric are highlighted in bold.

Most parameter efficient fine-tuning methods achieve better performance compared to linear-probe, as shown in Table I. This indicates that adjusting the features can effectively improve the accuracy of change detection tasks. The reason is that the features for upstream and downstream tasks should be different due to the differences in task and data domains. On the evaluation metric of IoU, we achieve an improvement of approximately 2%, 4%, and 3% compared to linear-probe on the WHU-CD, LEVIR-CD+, and CDD benchmarks, respectively. In our proposed change detection framework, even if only the linear layer is trained, we find that features from natural image pretrained models can also perform well for change detection tasks. Besides, all experiments were conducted on a single RTX3090 device, demonstrating that parameter efficient fine-tuning methods can reduce resource and training time requirements when applied to change detection.

Table I. Parameter Tuning methods on three benchmarks.

| Method | WHU-CD | | | LEVIR-CD+ | | | CDD | | |
|---|---|---|---|---|---|---|---|---|---|
| | F1 | IoU | OA | F1 | IoU | OA | F1 | IoU | OA |
| Linear-probe | 91.40 | 84.17 | 99.35 | 81.08 | 68.18 | 98.58 | 90.38 | 82.44 | 97.64 |
| Prefix-tuning | 90.56 | 82.75 | 97.69 | 81.63 | 68.96 | 98.62 | 90.25 | 82.23 | 97.62 |
| Lora | 91.84 | 84.91 | 99.37 | 84.22 | 72.74 | 98.77 | 91.39 | 84.14 | 97.88 |
| IA3 | 91.91 | 85.03 | 99.38 | 83.92 | 72.30 | 98.75 | 91.78 | 84.81 | 97.97 |
| Adapter | **92.53** | **86.11** | **99.43** | **84.49** | **72.98** | **98.78** | **92.81** | **85.59** | **98.22** |

Table II. Evaluation with SOTA methods on six benchmarks (Color convention: best, 2nd-best, and **3rd-best**).

| Method | LEVIR-CD | | | WHU-CD | | | DSIFN | | | CDD | | | S2Looking | LEVIR-CD+ |
|---|---|---|---|---|---|---|---|---|---|---|---|---|---|---|
| | F1 | IoU | OA | F1 | IoU | OA | F1 | IoU | OA | F1 | IoU | OA | F1 | F1 |
| FC-EF | 83.40 | 71.53 | 98.39 | 76.73 | 62.24 | 98.31 | 61.09 | 43.98 | 88.59 | 66.93 | 50.30 | 93.28 | 7.65 | 66.48 |
| FC-SC | 83.69 | 75.92 | 98.67 | 78.95 | 65.23 | 99.48 | 62.54 | 45.50 | 86.63 | 70.61 | 54.57 | 94.33 | 13.54 | 72.97 |
| FC-SD | 86.31 | 71.96 | 98.49 | 79.85 | 66.46 | 98.50 | 59.71 | 42.56 | 87.57 | 75.11 | 60.14 | 94.95 | 13.19 | 73.48 |
| DTC-SCN | 87.67 | 78.05 | 98.77 | **91.43** | **84.21** | **99.35** | 63.72 | 46.76 | 84.91 | 92.09 | 85.34 | 98.16 | 57.27 | 77.60 |
| STANet | 87.26 | 77.40 | 98.66 | 82.32 | 69.95 | 98.52 | 64.56 | 47.66 | 88.49 | 84.12 | 72.22 | 96.13 | 45.97 | 79.31 |
| IFNet | 88.13 | 78.77 | 98.87 | 83.40 | 71.52 | 98.83 | 60.10 | 42.96 | 87.83 | 84.00 | 71.91 | 96.03 | | |
| SNUNet | 88.16 | 78.83 | 98.82 | 83.50 | 71.67 | 98.71 | 66.18 | 49.45 | 87.34 | 83.89 | 72.11 | 96.23 | | |
| BIT | **89.31** | **80.68** | **98.92** | 90.53 | 83.39 | 99.34 | 69.26 | 52.97 | 89.41 | 88.90 | 80.01 | 97.47 | 61.85 | **82.80** |
| ChangeFormer | 90.40 | 82.48 | 99.04 | 88.57 | 79.49 | 99.12 | | | | 94.63 | 89.80 | 98.74 | **63.39** | |
| DINO-S | 88.47 | 79.33 | 98.88 | 92.53 | 86.11 | 99.43 | 72.91 | 57.37 | 91.51 | 92.81 | 86.59 | 98.22 | 64.09 | 84.49 |
| DINOv2-S | 89.55 | 81.08 | 98.98 | 92.38 | 85.84 | 99.40 | 68.50 | 52.09 | 89.39 | 93.93 | 88.55 | 98.49 | 67.29 | 85.66 |

*D. Evaluation with SOTA methods*

This section evaluates our strategy with state-of-the-art change detection methods on six change detection benchmarks. The experiments consider the DINO ViT-small and DINOv2 ViT-small as our pretrained model, which are named DINO-S and DINOv2-S in Table II. DINOv2 expands on the basis of DINO by increasing the model scale and using larger datasets in the context of large-scale models. Table II present the comparison with the state-of-the-art change detection models on six benchmarks. We chose FC-Net [4], DTC-SCN [50], STANet [46], IFNet [47], SNUNet [51], BIT [6], and ChangeFormer [7] as the state-of-the-art baseline models for comparison. The results on six datasets are illustrated in Table II. The values of each evaluation metric ranked best, second, third are highlighted in red, blue, and dark bold, respectively.

Based on DINO-S and DINOv2-S, our method achieves competitive or better performance than most prior arts. In change detection tasks, the F1 and IoU metrics are the most important evaluation metrics. However, previous methods have not reported IoU results on the S2Looking and LEVIR-CD+ datasets. On the evaluation metric of F1 score, we achieve the improvement approximately by 2%, 3%, 4% and 3% on WHU-CD, DSIFN, S2Looking and LEVIR-CD+ benchmarks compared with existing state-of-the-art methods. On other datasets, we are also able to achieve good performance with a difference in accuracy of less than 1%.

Note that ChangeFormer fully tunes a segmentation pretrained model based on vision transformer, taking more than 40 hours to train on a single RTX 8000 with a memory capacity of 48G in ChangeFormer. While our method simply adapts the pretrained vision transformer by training a small number of additional layers to change detection, taking half an hour to train on a single RTX3090 with a memory footprint of 9GB. The results show that our method is significantly more efficient than full fine-tuning a change detection model. This can also show that the proposed change detection framework is more general and can be loaded with more pretrained weights.

However, our method falls behind ChangeFormer on LEVIR-CD and CDD datasets. One reason is that LEVIR-CD is the smallest dataset among these six benchmarks and then the model may be overfitting. On some large datasets, e.g., S2Looking and DSIFN, our method shows large improvement. CDD is a dataset with seasonal variations, which may result in pseudo changes caused by seasonal factors. One potential explanation for this is that our experiment only employed simple data augmentation on the spatial scale, including horizontal and vertical flipping.

*E. Ablation studies*

In this section, we ablate tuning methods to study what properties make for a good tuning strategy and observe several intriguing properties.

**Ablation on Adapter.** As stated in Section III.C, the adapter layer is a bottleneck layer. The bottleneck dimension controls the number of introduced parameters by adapter layer. Lower bottleneck dimensions introduce fewer parameters at a possible cost to performance. We conducted an ablation study on the bottleneck feature dimension to investigate this effect on LEVIR-CD+ dataset. We denote the factor $r$ as the ratio of the feature dimension to the bottleneck dimension. As shown in Table III, the accuracy consistently improves when the factor $r$ increases up to 6 and reaches the saturation point when the bottleneck dimension is about 64. Since we set DINO-S as our default backbone, the feature dimension is 384.

The scaling factor $s$ is introduced to balance the task-agnostic features (generated by the original frozen branch) and the task-specific features (generated by the tunable bottleneck branch). We evaluate the adapter layer with multiple $s$ values and the results are also summarized in Table III. We found that it achieves optimal performance with $s = 1.0$. A larger or smaller $s$ value results in a slight drop in performance. Thus, we choose $s = 1.0$ as the default setting.

Table III. Factor study of Adapter on LEVIR-CD+ dataset.

| Factor | Factor Value | F1 | IoU |
|---|---|---|---|
| Feature factor $r$ | 4 | 84.46 | 73.10 |
| | 6 | **84.49** | **73.14** |
| | 8 | 83.28 | 71.34 |
| Scale factor $s$ | 0.1 | 82.08 | 69.60 |
| | 0.5 | 83.94 | 72.32 |
| | 1.0 | **84.49** | **73.14** |
| | 2.0 | 83.67 | 71.92 |
| | 4.0 | 83.75 | 72.04 |
| | Learned | 84.01 | 72.43 |

**Ablation on Lora.** The aim of Lora is to add extra layers to replace the behavior of updating parameters when do full fine-tuning. We insert linear layers in the self-attention layer as similar as Lora. As stated in section III.A, there are four linear transformation layers in the self-attention layer, namely $q$, $k$, $v$, and $o$ transformations. We denote $W_q, W_k, W_v, W_o$ as the reparameterization position of $q$, $k$, $v$, and $o$ weight matrix. We ablate Lora on the position and the low rank dim to study these effects on LEVIR-CD+ dataset. As shown in Table IV, using all $W_q, W_k, W_v, W_o$ layers can achieve the best performance. This shows that adjusting more parameters of the pretrained model can result in higher performance for downstream tasks such as change detection. Besides, when the low rank dimension is set to 1, it achieves the best performance and also reduces the tunable parameters and training time. One possible reason is that the change detection task is relatively simple and the amount of data is too small. When fine-tuning the model, many feature dimensions are redundant and only a few features are beneficial. Therefore, an extremely low-rank layer can achieve good results.

Table IV. Weight type study of Lora on LEVIR-CD+ dataset.

| Weight type | $r=1$ | $r=2$ | $r=4$ |
|---|---|---|---|
| $W_q$ | 83.29 | 79.61 | 80.94 |
| $W_q, W_v$ | 83.13 | 81.56 | 82.88 |
| $W_q, W_k, W_v, W_o$ | **84.22** | 84.03 | 83.61 |

**Scale the model size.** We further conduct experiments to compare the accuracy between DINO and DINOv2 when scaling the model size on four benchmarks. The ViT-base is chosen as backbone. We report F1 and IoU metric in Table V. Compared to DINO-S, DINO-B can achieve better F1 and IoU performance on the LEVIR-CD, WHU-CD, and CDD datasets. DINOv2-B can achieve better F1 and IoU performance on the LEVIR-CD, WHU-CD, and DSIFN datasets. DINO does well in DSIFN and CDD dataset. Compared with ViT-small, we observe that increasing the number of parameters in the ViT-base model can improve the performance of DINO and DINOv2. However, in most change detection datasets, ViT-small can achieve relatively good performance while requiring less training time and resource consumption. Therefore, we choose ViT-small as our default option.

Table V. Different model sizes on four benchmarks.

| | Method | LEVIR-CD | WHU-CD | DSIFN | CDD |
|---|---|---|---|---|---|
| F1 | DINO-S | 88.47 | 92.53 | **72.91** | 92.81 |
| | DINO-B | 88.96 | 92.98 | 71.19 | **94.07** |
| | DINOv2-S | 89.55 | 92.38 | 68.50 | 93.93 |
| | DINOv2-B | **89.90** | **93.77** | 70.58 | 93.58 |
| IoU | DINO-S | 79.33 | 86.11 | **57.37** | 86.59 |
| | DINO-B | 80.13 | 86.89 | 54.53 | **88.81** |
| | DINOv2-S | 81.08 | 85.84 | 52.09 | 88.55 |
| | DINOv2-B | **81.66** | **88.27** | 54.53 | 87.94 |

**Different pretrained models.** Here, we demonstrate the effectiveness of change detection framework with adapter layer on different pre-trained models, including SUP [8], DINO [29], MAE [34], BEIT-v2 [52]. In these methods, SUP refers to supervised training of the transformer on ImageNet, while the other methods are well-known self-supervised learning approaches on ImageNet. The change detection framework is performed on the LEVIR-CD+ dataset. In Table VI, we show the framework based on ViT-B backbone because most methods have a base version. DINO achieves the best performance and then we set it as our default choice in our experiments. Meanwhile, this also indicates that our method is compatible with most vision transformer-based methods, demonstrating the versatility of our approach and enabling it to benefit from more powerful image pretrained models.

Table VI. Different pretrained models on LEVIR-CD+ dataset.

| Arch. | Method | Pretrain data | F1 | IoU | OA |
|---|---|---|---|---|---|
| ViT-B/16 | SUP | IN-1K | 83.30 | 71.38 | 98.73 |
| | DINO | IN-1K | **85.22** | **74.24** | **98.84** |
| | MAE | IN-1K | 80.70 | 67.64 | 98.51 |
| | BEIT-v2 | IN-1K | 66.07 | 49.33 | 97.34 |

## V. CONCLUSION

In this work, we propose a new way to efficiently transfer pretrained image models for remote sensing change detection task. We compare the main efficient tuning methods and find that adapter layer is the best. Since only new added parameters are updated, the training cost is substantially lower than previous transformer-based change detection method. And we also achieve competitive and better performance on six benchmarks. The proposed framework is simple enough and it can also scale to different pretrained image models and multi-temporal change detection task. Therefore, in the future work, we might be able to detect object changes guided by text [53] instead of using one modal pretrained weight. And more temporal images will be also used to verify the generality of the proposed framework.